\documentclass[12pt]{article}

\usepackage{amsmath, newtxtext,newtxmath}

\usepackage{booktabs, bbm, amsthm}

\usepackage{graphicx}

\usepackage[letterpaper,margin=1in]{geometry}

\renewenvironment{abstract}
	{\quotation}
	{\endquotation}

\date{}

\makeatletter
\renewcommand{\fnum@figure}{\textbf{Figure \thefigure}}
\renewcommand{\fnum@table}{\textbf{Table \thetable}}
\makeatother

\usepackage{scicite}

\usepackage{url}
\usepackage{xspace}

\newcommand{\labelsupsub}[2]{\ensuremath{\ell^{(#1)}_{#2}}\xspace}
\newcommand{\indij}[2]{\ensuremath{\mathbbm{1}(#1 = #2)}\xspace}
\newcommand{\indand}[2]{\ensuremath{\mathbbm{1}(#1 \land #2)}\xspace}

\newcommand{\ellqtrue}{\labelsupsub{\text{true}}{q}}
\newcommand{\ellqi}{\labelsupsub{i}{q}}
\newcommand{\ellqj}{\labelsupsub{j}{q}}
\newcommand{\ellqk}{\labelsupsub{k}{q}}

\newcommand{\phat}[0]{\ensuremath{\hat{P}}}

\newcommand{\lbla}[0]{\ensuremath{\mathcal{A}}\xspace}
\newcommand{\lblb}[0]{\ensuremath{\mathcal{B}}\xspace}

\newcommand{\pa}[0]{\ensuremath{\phat_\lbla}\xspace}
\newcommand{\pb}[0]{\ensuremath{\phat_\lblb}\xspace}

\newcommand{\fliljlk}[3]{\ensuremath{f_{#1,#2,#3}}}
\newcommand{\pia}[1]{\ensuremath{\phat_{#1,\lbla}}}
\newcommand{\pib}[1]{\ensuremath{\phat_{#1,\lblb}}}
\newcommand{\pil}[1]{\ensuremath{\phat_{#1,\ell}}\xspace}

\newcommand{\svbar}[0]{\ensuremath{\; | \;}}

\newcommand{\prodDeltas}{\ensuremath{4 \, \Delta_{i,j} \, \Delta_{i,k} \, \Delta_{j,k}}}

\newcommand{\seconddiff}{\ensuremath{\Delta_{i,j,k}}}

\newcommand{\aeprev}{\ensuremath{\frac{1}{2}\left(1 -  \frac{(\seconddiff)}{\sqrt{\prodDeltas + (\seconddiff})^2}\right)}}

\newcommand{\corrijl}[3]{\ensuremath{\Gamma_{#1,#2}^{(#3)}}\xspace}
\newcommand{\corrijkl}[4]{\ensuremath{\Gamma_{#1,#2,#3}^{(#4)}}\xspace}

\newcommand{\mvsup}[1]{\ensuremath{#1^{(\text{MV})}}\xspace}
\newcommand{\aesup}[1]{\ensuremath{#1^{(\text{AE})}}\xspace}

\newcommand{\eqspace}[0]{\;\;\;\;}

\def\scititle{
	Algebraic Evaluation Theorems
}
\title{\bfseries \boldmath \scititle}

\author{
	Andr\'es Corrada-Emmanuel$^{1\ast\dagger}$,
	\small$^{1}$Data Engines, Easthampton 01027, USA.\\
	\small$^\ast$Corresponding author. Email: andres.corrada@dataengines.com
}

\begin{document} 

\newtheorem{theorem}{Theorem}[section]
\newtheorem{lemma}[theorem]{Lemma}

\maketitle

\begin{abstract} \bfseries \boldmath
Majority voting (MV) is the prototypical ``wisdom of the crowd''
algorithm. Theorems considering when MV is optimal
for group decisions date back to Condorcet's
1785 jury \emph{decision} theorem. The same error independence
assumption underlying the theorem can be used to prove
a jury \emph{evaluation} theorem that does purely algebraic
evaluation (AE) of juror performance based on a batch of
their decisions.
Three or more binary jurors
are enough to obtain the only two possible statistics of their
correctness on a test they took.
AE is superior to MV in three ways. First, its empirical assumptions
are looser and can handle jurors less than 50\% accurate in making
decisions. Second, it has point-like precision in evaluating
them given its assumption of error independence. This precision enables 
a multi-accuracy approach that has
higher labeling accuracy than MV and comes with empirical uncertainty bounds.
And, third, it is self-alarming about the failure of its error
independence assumption. Experiments using demographic data from
the American Community Survey confirm the practical utility
of AE over MV. Two implications of the theorem for AI safety are
discussed - a principled way to terminate infinite monitoring
chains (who grades the graders?) and the super-alignment problem 
(how do we evaluate agents doing tasks we do not understand?).
\end{abstract}

\noindent
A group of jurors are assembled to decide a series of cases
or questions. Each juror records their answers to the questions
on a computer form for multiple-choice exams ("bubble" sheet),
thereby erasing all semantic information about the questions
or their answers. Can we grade each juror using only these
forms? Evaluating experts while ignorant of the correctness of
their decisions is a topic of research in the safe development
and use of AI systems\cite{Burns2023, Khan2024, McAleese2024, zhuge2024agent}. 
This is an ancient and cross-disciplinary problem exemplified by
the Ship of Fools Allegory in Plato's 
\emph{The Republic} (Book V, 488b)\cite{Griffith2000}
A ship owner ignorant of sailing must select a member of the
crew that can captain the ship safely.
This mise-en-sc\`ene illustrates the economical, safety engineering, and
epistemological problems associated with evaluating experts when we,
ourselves, are ignorant of their degree of expertise.
The ship owner is the principal\cite{Laffont2002} that must delegate wisely to the
sailor agents while being ignorant of sailing.

Here we focus on the safety engineering aspect of monitoring experts without
ground truth -- unsupervised evaluation. Instead of giving 
the sailors a one question exam about who should be the captain, the owner 
could give the sailors a multiple-choice exam on sailing.
The evaluation theorems introduced here can be used to grade multiple
choice exams whenever experts make independent errors on them.
They are universally applicable, just like Majority Voting (MV).
Their only inputs are the integer counts of observed 
\emph{decision tuples} - the many ways they can agree and 
disagree when we align their responses question by question.

Plato dismissed MV as a mechanism by which sailors could self-evaluate
and select who was qualified to command the ship safely. This Report
agrees with Plato's dismissal of MV but not of ensembles. Three or
more error independent jurors taking binary multiple
choice exams can be graded perfectly (Supplementary Text). This approach
is fully algebraic, it uses no probability theory. Instead, it
considers the algebraic relations that must exist between the counts
of decisions tuples by jurors and statistics of their correctness
on a test.

Consider the decision tuples for three jurors. 
There are eight of them in a binary test ($2^3=8).$ Denote the
two labels by \lbla and \lblb. 
For example, a tuple like $(\lbla_i, \lblb_j, \lbla_k)$ or
any of the other seven
must be equal to the sum of an integer partition
by correct response. For tuple $(\lbla_i, \lblb_j, \lbla_k)$ its count,
$N_{\lbla_i, \lblb_j, \lbla_k}$, can be expressed as,
\begin{equation}
    N_{\lbla_i, \lblb_j, \lbla_k} = N_{\lbla_i, \lblb_j, \lbla_k; \lbla} +
        N_{\lbla_i, \lblb_j, \lbla_k; \lblb},
    \label{eq:by-true-label-partition}
\end{equation}
with $N_{\lbla_i, \lblb_j, \lbla_k; \lbla} \geq 0$ and 
$N_{\lbla_i, \lblb_j, \lbla_k; \lblb} \geq 0.$
The assumption of error independence allows us to rewrite this as,
\begin{equation}
    f_{\lbla_i, \lblb_j, \lbla_j} = \pa \pia{i} (1- \pia{j}) \pia{k} +
        (1 - \pa) (1 - \pib{i}) \pib{j} (1 - \pib{k}).
    \label{eq:aba-generating-polynomial}
\end{equation}
On the left side of Equation~\ref{eq:aba-generating-polynomial} we 
have the percentage of questions that the jurors answered with this
decision tuple, $f_{\lbla_i, \lblb_j, \lbla_j} = N_{\lbla_i, \lblb_j, \lbla_j}/Q.$
On the right side we have unknown statistics about the test and the correctness
of the jurors. The percentage, or prevalence, of questions correctly answered by
responses \lbla and \lblb are given by $\pa = Q_\lbla/Q$ and $\pb = Q_\lblb/Q$.
These are statistics of the answer key. Statistics about the average performance
of the classifier are given by the quantities $\pia{i}$ and $\pib{i}$ for a classifier $i.$ 
They also are integer ratios of the
form $N_{\lbla_i; \lbla}/Q_\lbla$ and $N_{\lblb_i; \lblb}/Q_\lblb$, respectively.

If we knew the values of these rational numbers, we would be able to grade
the average performance of the jurors. There are two interpretations
of these statistics of correctness. In classification tasks there
is a semantic equality between label responses across questions. AE acts like
a data streaming algorithm that uses the rational numbers 
$\{N_{\lbla_i, \lblb_j, \lbla_k}/Q\}$ to estimate statistical properties about the
classified items and how well the jurors did classifying them: \pa, \pb,
$\{\pia{i}\}_{i \in (1,2,3)}$, and $\{\pib{i}\}_{i \in (1,2,3)}.$ In multiple
choice exams, the labels across questions need not have any semantic connection.
Statistics like \pa and \pb have no meaning outside the test. In either case, 
we can calculate an average number of correct responses by juror $i$ as, 
\begin{equation}
    g_i = \pa \pia{i} + \pb \pib{i}.
\end{equation}

For three jurors, each of the eight possible decision tuples during a test
has a quartic polynomial similar to Equation~\ref{eq:aba-generating-polynomial}.
All of these polynomials
are expressible with just the prevalences \pa and \pb, and the label accuracies
$\{\pia{i}\}_{i \in (1,2,3)}$, and $\{\pib{i}\}_{i \in (1,2,3)}.$
The collection of eight polynomials can be viewed as setting up
an encode and decode framework for evaluation and hence the use of the term
Algebraic Evaluation (AE) to denote it. Given the assumption of error independence,
these polynomials ``encode" the frequency of decision tuples we will observe
in a test given the prevalences of question types and how correctly jurors
answered them. The ``decode" step, going from those observed frequencies of decision
tuples by the jurors to their accuracies is a problem in algebraic geometry
(Supplemental Text).

Decoding the observed frequencies of decision tuples leads to two solutions
for the evaluation statistics. This ambiguity is also implicit in the use
of MV. The majority vote is not always right. So for some questions,
the minority vote is the correct decision. Condorcet's theorem breaks
the ambiguity with the assumption that all the jurors must be better than
50\% correct at answering \lbla and \lblb type questions to justify
MV as the choice that minimizes \emph{total} labeling errors.
AE only has the assumption of error
independence. Given that assumption, one of the two solutions returned by
the decoding step is the true evaluation of the jurors. There are
many principled ways to resolve this ambiguity. The one used in the experiments
reported here is to pick the solution with greatest total label accuracy,
\begin{equation}
    \sum_{i \in (1,2,3)} (\pia{i}^{\text{(sol)}} + \pib{i}^{\text{(sol)}}),
\end{equation}
a looser assumption than Condorcet's theorem that can accommodate jurors
less than 50\% accurate on either label.

AE improves over MV as a starting point for decisions also. Consider
experiments carried out with classifiers labeling demographic datasets from the 
\textsc{folktables} project\cite{folktables}. These are demographic records,
for individuals that participated in the \emph{American Community
Survey} carried out annually by the US Census. Four neural networks with
identical architecture where trained on disjoint features and data from a small
subset (60K) of about 1.5M 2018 records\cite{methods}. This made them nearly
error independent ($\approx0.02$). Using AE, we can construct the integer partition
of the decision tuples of an ensemble by true answer type. 
Table~\ref{tab:by-true-label-partition} shows one table from labeling
20K records as to whether the individual was employed or not. Three
of the decision tuples, (\lbla, \lbla, \lblb), (\lbla, \lblb, \lbla),
and (\lblb, \lbla, \lbla), would lead to less labeling errors if we
opted for the minority vote, \lblb. This is an empirical version of the
concept of multi-accuracy or No-Access-OI outcome 
indistinguishability\cite{Dwork2020}
in discussions of algorithmic fairness. These improved labeling
decisions from AE come with empirical bounds of their error since
we are also estimating how many have the opposite label.
This makes AE analogous
to conformal prediction\cite{Vovk2005}, but instead of giving uncertainty bounds based
on internal scores of a single classifier it gives them for ensembles of
three or more classifiers based on their decision tuples. Note that MV
always yields the uninformative estimate of 0 for its own errors.

Looser assumptions, better as a starting point for evaluation and decisions,
the last advantage of AE over MV is its ability to warn that the
assumption of independent test errors is not correct. The warning
is, itself, not perfect. It has false positives - some ensembles will be
correlated but AE will return rational estimates for the grades. But
no false negatives - if AE returns irrational numbers, the classifiers
did not act error independently on the test (Supplemental Text). The prevalence
estimates for label \lbla given by MV and AE show the algebraic nature
of the AE alarm. The MV estimate of \pa is,
\begin{equation}
    \mvsup{\pa} = \fliljlk{\lbla}{\lbla}{\lbla} + \fliljlk{\lbla}{\lbla}{\lblb} +
    \fliljlk{\lbla}{\lblb}{\lbla} + \fliljlk{\lblb}{\lbla}{\lbla}
    \label{eq:mv-a-prev}
\end{equation}
and thus, always a rational number estimate since it is a sum of rational numbers.
In contrast, the AE estimate of label \lbla prevalence comes from solving
a quadratic polynomial and involves a square root,
\begin{equation}
   \aesup{\pa} = \aeprev,
   \label{eq:ae-a-prev}
\end{equation}
where the deltas ($\Delta_{i,j,k}, \Delta_{i,j}, \Delta{i,k}, \Delta_{j,k}$) 
are polynomial moments of the decision tuple frequencies\cite{methods}.
As polynomials of rational numbers, the deltas are rational numbers.
But there is no guarantee that the square root on the RHS of Equation~\ref{eq:ae-a-prev}
will be rational. And for most correlated ensembles it is not (Supplemental Text).
Algebraic numbers partly certify their correctness.

Grading experts without answer keys can mitigate two epistemological problems
in AI safety -- infinite monitoring chains and super-alignment. Any work on
monitoring AI agents with other agents raises the question of what verifies
the monitors. We have no framework to encompass all the theories experts
can have about the World and still be correct.
This makes proposals for safe guaranteed AI\cite{Dalrymple2024}
with single agents difficult to implement. AE is empirical so it bypasses
this lack of knowledge. It has finite and complete
models for the results of tests experts take, not about the theories those
experts used to make the decisions. It can cap monitoring chains after one step. 
Experts that assess experts can be viewed as taking a test about their grading.
In such cases, AE can be used to grade the assessors as in the DevAI\cite{zhuge2024agent}
evaluation framework proposed for agents developing AI apps.
It consists of 55 AI app development tasks that are evaluated across 365
hierarchical requirements so as to emphasize step-by-step problem solving
ability. Each of these requirements is graded as "Completed" or not by
human or AI agents\cite{zhuge2024agent}, a binary testing task.
The same lack of knowledge occurs in the problem of super-alignment.
As AI agents carry out tasks or produce knowledge we
cannot understand, how can we supervise them? Can weak supervisors
control agents smarter than them?\cite{Burns2023}
Researchers in super-alignment have commented on the asymmetry
between evaluating experts and being 
one.\cite{Burns2023,McAleese2024,Khan2024} AE exhibits this
asymmetry and could be a useful tool for monitoring agents
when us, the principals, are ignorant.






\begin{table} 
	\centering
	\caption{\textbf{Partition, by true label, of the observed decision counts
    of three binary classifiers labeling a 20K dataset from the 2018
    American Community Survey.} Algebraic evaluation (AE) is compared with
    majority voting (MV). Their estimates can be compared with the actual or
    ground truth partition. Algebraic evaluation is a better estimator
    of classifier labeling accuracy and leads to less labeling errors
    than majority voting.}
	\label{tab:by-true-label-partition} 
	\begin{tabular}{llcccccc}
    & \multicolumn{1}{c}{method}  & \multicolumn{2}{c}{AE} & \multicolumn{2}{c}{actual} & \multicolumn{2}{c}{MV}
    \\\cmidrule(lr){3-4}\cmidrule(lr){5-6}\cmidrule(lr){7-8}
    & \multicolumn{1}{c}{true label} & \lbla  & \lblb & \lbla  & \lblb & \lbla  & \lblb \\\midrule
    decisions & observed count & \multicolumn{6}{c}{}\\
    (\lbla, \lbla, \lbla) & 568 & 399  & 169 & 424 & 144 & 568 & 0\\
    (\lbla, \lbla, \lblb) & 553 & 133  & 420 & 168 & 385 & 553 & 0 \\
    (\lbla, \lblb, \lbla) & 649 & 253  & 396 & 283 & 366 & 649 & 0\\
    (\lblb, \lbla, \lbla) & 1813 & 416 & 1397 & 415 & 1398 & 1813 & 0\\
    (\lblb, \lblb, \lbla) & 3534 & 264 & 3270 & 252 & 3282 & 0 & 3534\\
    (\lblb, \lbla, \lblb) & 3607 & 139 & 3468 & 194 & 3413 & 0 & 3607\\
    (\lbla, \lblb, \lblb) & 1068 & 84  & 984 & 129 & 939 & 0 & 1068\\
    (\lblb, \lblb, \lblb) & 8208 & 88  & 8120 & 135 & 8073 & 0 & 8208
    \end{tabular}
\end{table}


\clearpage 

%
\bibliography{science_template} 
\bibliographystyle{sciencemag}

%
%
%
%
%
%


\section*{Acknowledgments}
I would like to thank Ilya Parker for fruitful discussions on
the topic of unsupervised evaluation and its role in AI safety.
\paragraph*{Funding:}
This work was partly supported by a 2024 Summer Research Fellowship with the U.S.\ Naval Research Laboratory.
\paragraph*{Author contributions:}
A.\ Corrada-Emmanuel is the sole author of this paper.
\paragraph*{Competing interests:}
This jury evaluation theorem was first published as one of
the main claims in US Patent XXXXXX in 2010. In hindsight, the
patent was improperly granted and its owner, Data Engines, is
allowing it to expire.

\paragraph*{Data and materials availability:}



\subsection*{Supplementary materials}
Materials and Methods\\
Supplementary Text\\
Data S1\\
Data S2\\
Data S3\\
Data S4\\
Code S1\\
Code S2


\newpage


\renewcommand{\thefigure}{S\arabic{figure}}
\renewcommand{\thetable}{S\arabic{table}}
\renewcommand{\theequation}{S\arabic{equation}}
\renewcommand{\thepage}{S\arabic{page}}
\setcounter{figure}{0}
\setcounter{table}{0}
\setcounter{equation}{0}
\setcounter{page}{1} 


\begin{center}
\section*{Supplementary Materials for\\ \scititle}

Andr\'es~Corrada-Emmanuel$^{\ast\dagger}$\\
\small$^\ast$Corresponding author. Email: andres.corrada@dataengines.com\\
\end{center}

\subsubsection*{This PDF file includes:}
Materials and Methods\\
Supplementary Text\\
Captions for Code S1 to S3\\
Captions for Data S1 to S4

\subsubsection*{Other Supplementary Materials for this manuscript:}
Code S1 to S3\\
Data S1 to S2

\newpage


\subsection*{Materials and Methods}

Experiments were carried out using 2018 demographic data
from the US Census American Community Survey as surfaced by the 
\textsc{folktables} project\cite{ding2021retiring}. 
It contains demographic attributes
for 3,236,107 individuals.
Various datasets and tasks are defined in \textsc{folktables}
to promote common benchmarks. The one used here is the \textsc{ACSEmployment} 
dataset associated with the binary task of labeling records as either
"does-not-have-employment" or 
"has-employment".
There are 1,763,385 records of the first label (non-employed), 
and 1,472,722 of the second (employed).

Each record contains 16 features associated with demographic
categories. These are, somewhat self-descriptive - AGEP, SCHL, MAR,
RELP, DIS, ESP, CIT, MIG, MIL, ANC, NATIVITY, DEAR, DEYE, DREM, SEX, RAC1P. 
Detailed descriptions of their definitions
can be found in the PUMS Data Dictionaries\cite{PUMS}. Of these 16 attributes,
the sensitive attributes AGEP, SEX, RAC1P (age, gender, and race) were held out 
because of their importance in the algorithmic fairness community. A small subset
of the experimental results focus on the multi-accuracy ability of AE by looking
at the performance of these models on datasets based on gender and race.
The remaining 13 features were available for all subsequent experiments
for predicting the employment status of individuals in the survey.

\subsubsection{Engineering nearly error independent classifiers}

The practical utility of the AE depends on how well we can actually produce classifiers
that are nearly error independent when tested. This is ultimately an engineering
problem. The theory of AE may be universal since it lacks probability assumptions
but its application is not. This is not unusual. Error-detecting and correcting
codes act similarly. There is the theory of, say, error-detection for bit flips
in an encoding. But the trade-off between partial detection of errors and speed
or memory means that we engineer codes for their application context.

There is nothing in the theory of AE that can guarantee that it will work.
The phenomena or features used to analyze it may be such that it is impossible
to train at least three error independent classifiers. The theory of AE may
be simple and thus widely applicable in principle, but the robustness of
its application is an open engineering problem that needs to be ascertained
on each use.

The goal of these experiments was to demonstrate that one can build ensembles
of nearly error-independent classifiers. The goal was not to produce the most
accurate classifiers for the \textsc{ACSEmployment} task. Two techniques were
used to induce the most error independence:
\begin{enumerate}
    \item Disjoint data.
    \item Disjoint features.
\end{enumerate}

Creating disjoint partitions of the data was simple given the large size of the 2018 
\textsc{ACSEmployment} dataset and the details are described below when data splits
are discussed. The 13 features available for can be split into
4 disjoint sets of 3 (leaving one out) in 200,200 possible ways. A random search
of about 100 of these choices was sufficient to find the partition that is used in
all the experiments reported here - ((SCHC, MIG, DREM), (MAR, CIT, DEAR), (RELP,
DIS, ANC), (MIL, NATIVITY, DEYE)).

A third way to induce error independence on tests would be to train each classifier
on different types of algorithms. This was not necessary in the case of the
\textsc{ACSEmployment} task. The four classifiers were trained using the
same Neural Network architecture. These are Fully Connected Architectures six 
layers deep and 9 nodes wide. Each contained 416 parameters. 
Details on the architecture and the
commands used in the Wolfram language to produce the classifiers can be found in
the Code Notebook (S2).

\subsubsection*{Training, validation, and testing splits}

The large size of the \textsc{ACSEmployment} dataset allows training and validation
dataset sizes that are large enough to get label accuracies around 65\% but small 
relative to the testing set. Training and validation was carried out on about 50K
records total but tested on random 20K samples from the 3M records remaining.
The classifiers were trained and validated only once. The experiments
focused on variation of AE and MV predictions given the same fixed
classifiers.

Four thousand evaluations were carried out with the held-out testing data (non-employed:
1,679,385, employed: 1,388,722). Each evaluation had 20K records to label. The evaluations
were carried out at four settings of the prevalences. The unemployed prevalence in a test,
\pa, was set to 2/10, 4/10, 6/10, and 8/10.

\subsubsection*{Evaluating a single 20K test}

The basic unit of the experiments was a comparison of AE and MV on a
randomly drawn dataset of size 20K.
For each such labeling test, we have the counts, by-true label, of
the decisions of the members of the ensemble. This is the ground truth of a single
evaluation. This is the integer partition that is illustrated in 
Table~\ref{tab:by-true-label-partition}.

In unsupervised settings, no one has access to these integer partitions.
The inputs for AE and MV are the sums of these ground truth counts - the
observable counts of decisions tuples by the ensemble. These inputs are
used differently by AE and MV depending on whether we are doing evaulation
or decision.

\subsubsection*{Evaluation formulas for AE and MV}

MV can be turned into an evaluation
algorithm by imputing an answer key for each item based on the majority decision for
it. The closed form formulas for the MV evaluations are,
\begin{align}
    \mvsup{\pa} & = \fliljlk{\lbla}{\lbla}{\lbla} + \fliljlk{\lbla}{\lbla}{\lblb} +
                    \fliljlk{\lbla}{\lblb}{\lbla} + \fliljlk{\lblb}{\lbla}{\lbla}\\
    \mvsup{\pb} & = \fliljlk{\lblb}{\lblb}{\lblb} + \fliljlk{\lblb}{\lblb}{\lbla} +
                    \fliljlk{\lblb}{\lbla}{\lblb} + \fliljlk{\lbla}{\lblb}{\lblb}\\
    \mvsup{\pia{i}} & = (\fliljlk{\lbla}{\lbla}{\lbla} + \fliljlk{\lbla}{\lbla}{\lblb} +
                    \fliljlk{\lbla}{\lblb}{\lbla}) / \mvsup{\pa}\\
    \mvsup{\pib{i}} & = (\fliljlk{\lblb}{\lblb}{\lblb} + \fliljlk{\lblb}{\lblb}{\lbla} +
                    \fliljlk{\lblb}{\lbla}{\lblb}) / \mvsup{\pb}\\
    \mvsup{\pia{j}} & = (\fliljlk{\lbla}{\lbla}{\lbla} + \fliljlk{\lbla}{\lbla}{\lblb} +
                     \fliljlk{\lblb}{\lbla}{\lbla}) / \mvsup{\pa}\\
    \mvsup{\pib{j}} & = (\fliljlk{\lblb}{\lblb}{\lblb} + \fliljlk{\lblb}{\lblb}{\lbla} +
                     \fliljlk{\lbla}{\lblb}{\lblb}) / \mvsup{\pb}\\
    \mvsup{\pia{k}} & = (\fliljlk{\lbla}{\lbla}{\lbla}  +
                    \fliljlk{\lbla}{\lblb}{\lbla} + \fliljlk{\lblb}{\lbla}{\lbla}) / \mvsup{\pa}\\
    \mvsup{\pib{k}} & = (\fliljlk{\lblb}{\lblb}{\lblb}  +
                    \fliljlk{\lblb}{\lbla}{\lblb} + \fliljlk{\lbla}{\lblb}{\lblb}) / \mvsup{\pb}
\end{align}

The evaluation estimates obtained from the AE can be obtained in a variety of ways. In the
accompanying Wolfram computational notebook (S1), the built-in polynomial system solver
function \textsc{Solve} was used. Any symbolic computational platform that can carry out Buchberger's
algorithm\cite{Buchberger} could also be used. Estimates can also be computed using the following
equations. The computation proceeds in two stages. In the first stage, the frequencies
of the decision tuples are used to compute polynomial moments of them, the Delta
expressions. These are then used to carry out a series of polynomial operations
starting with solving a quadratic equation for the label \lbla prevalence.

There are moments for single classifiers, the pairs, and the trio.
\begin{align}
    f_{\lblb_i} & = \fliljlk{\lblb}{\lbla}{\lbla} + \fliljlk{\lblb}{\lbla}{\lblb} +
        \fliljlk{\lblb}{\lblb}{\lbla} + \fliljlk{\lblb}{\lblb}{\lblb} \\
    f_{\lblb_j} & = \fliljlk{\lbla}{\lblb}{\lbla} + \fliljlk{\lbla}{\lblb}{\lblb} +
        \fliljlk{\lblb}{\lblb}{\lbla} + \fliljlk{\lblb}{\lblb}{\lblb} \\
    f_{\lblb_k} & = \fliljlk{\lbla}{\lbla}{\lblb} + \fliljlk{\lbla}{\lblb}{\lblb} +
        \fliljlk{\lblb}{\lbla}{\lblb} + \fliljlk{\lblb}{\lblb}{\lblb} \\
    \Delta_{i,j} & = (\fliljlk{\lblb}{\lblb}{\lbla} + \fliljlk{\lblb}{\lblb}{\lblb}) -
        f_{\lblb_i} f_{\lblb_j} \\
    \Delta_{i,k} & = (\fliljlk{\lblb}{\lbla}{\lblb} + \fliljlk{\lblb}{\lblb}{\lblb}) -
        f_{\lblb_i} f_{\lblb_k} \\
    \Delta_{j,k} & = (\fliljlk{\lbla}{\lblb}{\lblb} + \fliljlk{\lblb}{\lblb}{\lblb}) -
        f_{\lblb_j} f_{\lblb_k} \\
    \Delta_{i,j,k} & = \fliljlk{\lblb}{\lblb}{\lblb} -(f_{\lblb_i} f_{\lblb_j} f_{\lblb_k} + 
        f_{\lblb_i} \Delta_{j,k} + f_{\lblb_j} \Delta_{i,k} + f_{\lblb_k} \Delta_{i,j}).
\end{align}

Using these empirical quantities, the error independent evaluation estimates are,
\begin{enumerate}
    \item The quadratic polynomial for \pa,
    \begin{align}
        0 & = a \pa^2 + b \pa + c \\
        a & =  \Delta_{i,j,k}^2 +
            4 \Delta_{i,j} \Delta_{i,k} \Delta_{j,k} \\
        b & = -a \\
        c & = \Delta_{i,j} \Delta_{i,k} \Delta_{j,k}
    \end{align}

    \item The linear equation for \pia{i}, once a solution for \pa is selected,
    \begin{align}
        0 & = a + b \pa + c \pia{i} \\
        a & = \Delta_{i,j,k} 
        (\Delta_{i,j,k} - \Delta_{j,k}(1 - f_{\lblb_i}))  +
            2 \Delta_{i,j} \Delta_{i,k} \Delta_{j,k} \\
        b & = 4 \Delta_{i,j} \Delta_{i,k} \Delta_{j,k} -
             \Delta_{i,j,k}^2 \\
        c & = \Delta_{j,k} \Delta_{i,j,k}
    \end{align}

    \item The linear equation for \pib{i}, once a solution for \pa is selected,
    \begin{align}
        0 & = a + b \pa + c \pib{i} \\
        a & = f_{\lblb_i} \Delta_{j,k} \Delta_{i,j,k} -
            2 \Delta_{i,j} \Delta_{i,k} \Delta_{j,k} \\
        b & =  \Delta_{i,j,k}^2 +
            4 \Delta_{i,j} \Delta_{i,k} \Delta_{j,k} \\
        c & = - \Delta_{j,k} \Delta_{i,j,k}
    \end{align}

    \item The linear equation for \pia{j}, once a solution for \pa is selected,
    \begin{align}
        0 & = a + b \pa + c \pia{j} \\
        a & = \Delta_{i,j,k} 
        (\Delta_{i,j,k} - \Delta_{i,k}(1 - f_{\lblb_j}))  +
            2 \Delta_{i,j} \Delta_{i,k} \Delta_{j,k} \\
        b & = 4 \Delta_{i,j} \Delta_{i,k} \Delta_{j,k} -
            \Delta_{i,j,k}^2 \\
        c & = \Delta_{i,k} \Delta_{i,j,k}
    \end{align}

    \item The linear equation for \pib{j}, once a solution for \pa is selected,
    \begin{align}
        0 & = a + b \pa + c \pib{j} \\
        a & = f_{\lblb_j} \Delta_{i,k} \Delta_{i,j,k} -
            2 \Delta_{i,j} \Delta_{i,k} \Delta_{j,k} \\
        b & = \Delta_{i,j,k}^2 +
            4 \Delta_{i,j} \Delta_{i,k} \Delta_{j,k} \\
        c & = -\Delta_{i,k} \Delta_{i,j,k}
    \end{align}

    \item The linear equation for \pia{k}, once a solution for \pa is selected,
    \begin{align}
        0 & = a + b \pa + c \pia{k} \\
        a & =  \Delta_{i,j,k}
        (\Delta_{i,j,k} - \Delta_{i,k}(1 - f_{\lblb_j}))  +
            2 \Delta_{i,j} \Delta_{i,k} \Delta_{j,k} \\
        b & = 4 \Delta_{i,j} \Delta_{i,k} \Delta_{j,k} -
            \Delta_{i,j,k}^2 \\
        c & = \Delta_{i,j} \Delta_{i,j,k}
    \end{align}

    \item The linear equation for \pib{k}, once a solution for \pa is selected,
    \begin{align}
        0 & = a + b \pa + c \pib{k} \\
        a & = f_{\lblb_k} \Delta_{i,k} \Delta_{i,j,k} -
            2 \Delta_{i,j} \Delta_{i,k} \Delta_{j,k} \\
        b & = \Delta_{i,j,k}^2 +
            4 \Delta_{i,j} \Delta_{i,k} \Delta_{j,k} \\
        c & = - \Delta_{i,j} \Delta_{i,j,k}
    \end{align}
\end{enumerate}

\subsubsection*{The labeling decision task for AE and MV}

The decision task for the binary classifiers is to correctly label each item in a 20K test set.
The baseline for comparisons is the number of errors that would be made by using the ground
truth for the integer partition of each ensemble decision. Once we have an integer partition
of the decision tuples by either AE or MV, the label decision is the one that minimizes
the errors given its own estimates of the partition.

\subsubsection*{Error correlation measurements}

The practical applicability of the AE hinges on the ability to engineer
ensembles that are nearly error independent. As described above, disjoint
data and features were used in the experiments reported here. This section
details how it is calculated given the by-true label decision counts. In
general, all 4K evaluations had error correlations of the order of
$10^{-2}$. Fully correlated classifiers at 60\% label accuracy have a value
of $0.24$, so the experiments confirm that creating nearly error independent
classifiers is possible.

The test error correlation for a pair of binary classifiers is defined as,
\begin{equation}
        \corrijl{i}{j}{\ell}  = \frac{1}{Q_\ell} \sum_{q=1}^{Q} \left( \indij{\ellqi}{\ell} - \pil{i} \right) \left( \indij{\ellqj}{\ell} - \pil{j} \right) \indij{\ell}{\ellqtrue}
    \end{equation}
The definition for a trio is,
\begin{equation}
        \corrijkl{i}{j}{k}{\ell} = \frac{1}{Q_\ell} \sum_{q=1}^{Q} \left( \indij{\ellqi}{\ell} - \pil{i} \right) \left( \indij{\ellqj}{\ell} - \pil{j} \right)
        \left( \indij{\ellqk}{\ell} - \pil{k} \right) \indij{\ell}{\ellqtrue}.
    \end{equation}

These expressions allow us to define error-independence empirically as a point-value set of
expressions - all pair and trio correlations are zero for both labels,
\begin{align}
    \{\corrijl{i}{j}{\ell} = 0\}_{\ell \in (\lbla, \lblb)},\; & \;\{\corrijkl{i}{j}{k}{\ell} = 0\}_{\ell \in (\lbla, \lblb)}. \\
\end{align}

\newpage 


\subsection*{Supplementary Text}

Two main theorems are proved here. The first is the ``encode'' step of
Algebraic Evaluation (AE) - the set of polynomials that
generate the observed decision tuple counts given statistics of how correct
jurors were responding to the test. The second is the
``decode'' step - given the frequency of decision tuples observed in
the juror responses to the test,
what are the two possible juror evaluations? A third theorem uses
the first two to characterize the self-alarming properties of
AE.

\subsubsection{Models for ground truth and tests}

A juror $i$ produces a record of the form,
\begin{equation}
    \{ (q, \ell_i^{(q)} \}_{q=1}^{Q},
\end{equation}
for a multiple-choice test with $Q$ questions. The algebra of this
\emph{model} or representation of the test can have two interpretations.
When classifiers are tested, there is a semantic connection between
label responses to different items. The experiments reported here
are of this type. The label "has-employment", encoded as \lblb, means
the same thing for each of the labeled records. The second interpretation
corresponds to testing experts with questions that have a fixed number
of responses.

In either of the interpretations above, we can define notions of correctness
in the test by writing the ground truth for the correct answers in the same
representation. We assume there exists an \emph{answer key} for the test
that can be represented in the form,
\begin{equation}
    \{ \labelsupsub{\text{true}}{q}  \}_{q=1}^{Q}
    \label{eq:bubble-sheet-gt}
\end{equation}
This answer key allows us to formally define all the statistics of correctness
that are needed.

\subsubsection{Definitions of statistics of correctness for a test}

There are two types of sample statistics for these test models. The
first type are statistics about the type of questions on the test.
The integer version of those are the number of questions it contained for
each label,
\begin{equation}
    Q_\ell = \sum_{q=1}^{Q} \indij{\ell}{\labelsupsub{\text{true}}{q}},
\end{equation}
where \indij{\ell_i}{\ell_j} is the indicator function,
\begin{equation}
    \indij{\ell_i}{\ell_j} = \begin{cases}
        1 & \text{if } \ell_i = \ell_j\\
        0 & \text{otherwise}
    \end{cases}
\end{equation}
The rational number version of these statistics are the \emph{prevalences}
of each label in the test,
\begin{equation}
    \phat_\ell = Q_\ell / Q.
\end{equation}

The second type of test sample statistic we need are the label accuracies
for classifiers,
\begin{equation}
    \phat_{i,\ell} = \begin{cases}
        1/Q_\ell \sum_{q=1}^{Q} \indij{\ellqi}{\ellqtrue} \indij{\ell}{\ellqtrue} & \text{if } Q_\ell \neq 0\\
        0 & \text{otherwise}
    \end{cases}
\end{equation}

These two types of test statistics $\{\phat_\ell\}_{\ell \in \mathcal{L}}$ and 
$\{\phat_{i,\ell} \}_{i \in (1,2, \ldots), \ell in \mathcal{L}}$
will now be shown to be enough to generate all possible frequencies
of the decision tuple counts of three error independent binary classifiers.

\subsubsection*{The encoding theorem for error independent binary classifiers}

The first theorem proves that we can predict the frequencies of the agreements
and disagreements between three classifies given the test statistics defined
in the previous section.

\begin{theorem}
    Given the prevalence of question types $\{\phat_\ell\}_{\ell \in \mathcal{L}}$
    in a test of size $Q$,
    and the label accuracy of the three error independent classifiers, 
    $\{\phat_{c,\ell} \}^{c \in (i,j,k)}_{\ell \in (\lbla, \lblb)}$,
    the observed frequencies of their eight possible joint decisions,
    \begin{equation}
        f_{\ell_i, \ell_j, \ell_k} = \frac{n(\ell_i, \ell_j, \ell_k)}{Q},
    \end{equation}
    are given by the following polynomial set,
    \begin{flalign}
  \fliljlk{\lbla}{\lbla}{\lbla}  = \eqspace &  \pa \pia{i} \pia{j} \pia{k}& + \eqspace & \; \pb (1 - \pib{i}) (1 - \pib{j}) (1 - \pib{k})\\
  \fliljlk{\lbla}{\lbla}{\lblb}  = \eqspace &  \pa \pia{i} \pia{j} (1-\pia{k}) & + \eqspace & \; \pb (1 - \pib{i}) (1 - \pib{j}) \pib{k}\\
  \fliljlk{\lbla}{\lblb}{\lbla}  = \eqspace &  \pa \pia{i} (1 - \pia{j}) \pia{k} & + \eqspace & \; \pb (1 - \pib{i}) \pib{j} (1 - \pib{k})\\
  \fliljlk{\lblb}{\lbla}{\lbla}  = \eqspace &  \pa (1 - \pia{i}) \pia{j} \pia{k} & + \eqspace & \; \pb \pib{i} (1 - \pib{j}) (1 - \pib{k})\\
  \fliljlk{\lblb}{\lblb}{\lbla}  = \eqspace & \pa  (1 - \pia{i})  (1 - \pia{j}) \pia{k} & + \eqspace & \; \pb \pib{i} \pib{j} (1 - \pib{k})\\
  \fliljlk{\lblb}{\lbla}{\lblb}  = \eqspace & \pa  (1 - \pia{i})  \pia{j} (1 - \pia{k}) & + \eqspace & \; \pb \pib{i} (1 - \pib{j}) \pib{k}\\
  \fliljlk{\lbla}{\lblb}{\lblb}  = \eqspace & \pa  \pia{i}  (1 - \pia{j}) (1 - \pia{k}) & + \eqspace & \; \pb (1 - \pib{i}) \pib{j} \pib{k}\\
  \fliljlk{\lblb}{\lblb}{\lblb}  = \eqspace & \pa  (1 - \pia{i})  (1 - \pia{j}) (1 - \pia{k}) & + \eqspace & \; \pb \pib{i} \pib{j} \pib{k}.
\end{flalign}
\begin{proof}
    The existence of an answer key for the exam implies that the observed counts for any label decisions
    by the members of the trio must obey the following integer partition,
    \begin{equation}
        n(\ell_i, \ell_j, \ell_k) = n(\ell_i, \ell_j, \ell_k \svbar \lbla) + n(\ell_i, \ell_j, \ell_k \svbar \lblb),
    \end{equation}
    with $n(\ell_i, \ell_j, \ell_k \svbar \ell) \geq 0.$ We can write any of these conditional integer counts using
    our ground truth model and its associated indicator functions as follows,
    \begin{multline}
        n(\ell_i, \ell_j, \ell_k \svbar \ell) = \sum_{q=1}^{Q} \indij{\ell_i}{\labelsupsub{i}{q}} \indij{\ell_j}{\labelsupsub{j}{q}} \indij{\ell_k}{\labelsupsub{k}{q}} \indij{\ell}{\ellqtrue}\\
        = \sum_{q=1}^{Q} \indand{(\ell_i = \labelsupsub{i}{q})}{(\ellqtrue = \ell)} \indand{(\ell_i = \labelsupsub{i}{q})}{(\ellqtrue = \ell)} \indand{(\ell_i = \labelsupsub{i}{q})}{(\ellqtrue = \ell)}.
    \end{multline}
    The count of times the classifiers had decision pattern $(\ell_i, \ell_j, \ell_k \svbar \ell)$ is written as a sum of AND
    operations where each classifier must have responded with the required label in the pattern and the correct answer
    to question $q$ is indeed $\ell.$
    
    Each of the the indicator functions of the form, \indand{(\ell_i = \labelsupsub{i}{q})}{(\ellqtrue = \ell)},
    appears twice given the correct label, when it agrees with it and when it does not. In the case
    of binary classification this allows us to express the patterns where a classifier is incorrect with
    one minus its correct indicator function. For example, \lbla correct questions have the
    identity,
    \begin{equation}
        \indand{(\labelsupsub{(i}{q} = \lblb)}{(\ellqtrue = \lbla)} = 
        (1 - \indand{(\labelsupsub{(i}{q} = \lbla)}{(\ellqtrue = \lbla)}).
    \end{equation}
    This reduces the proof to verifying if we can write averages of indicator products
    as products of their averages. To establish this, we need a definition of correlation
    in terms of statistics of the indicators.

    The error correlation, \corrijl{i}{j}{\ell}, of two binary classifiers, i and j, given true label, $\ell$, is defined as,
    \begin{align}
        \corrijl{i}{j}{\ell} & = \frac{1}{Q_\ell} \sum_{q=1}^{Q} \left( \indij{\ellqi}{\ell} - \pil{i} \right) \left( \indij{\ellqj}{\ell} - \pil{j} \right) \indij{\ell}{\ellqtrue}\\
        & = \left( \frac{1}{Q_\ell} \sum_{q=1}^{Q} \indij{\ellqi}{\ell} \indij{\ellqj}{\ell} \indij{\ell}{\ellqtrue} \right) - \pil{i} \pil{j}.
    \end{align} 
    Under the assumption that this is zero for the test, we can write any average of the product of two indicators
    as the product of their averages,
    \begin{equation}
        \frac{1}{Q_\ell} \sum_{q=1}^{Q} \indij{\ellqi}{\ell} \indij{\ellqj}{\ell} \indij{\ell}{\ellqtrue} = \pil{i} \pil{j}.
    \end{equation}

    Similarly, we can define a 3-way correlation as,
    \begin{equation}
        \corrijkl{i}{j}{k}{\ell} = \frac{1}{Q_\ell} \sum_{q=1}^{Q} \left( \indij{\ellqi}{\ell} - \pil{i} \right) \left( \indij{\ellqj}{\ell} - \pil{j} \right)
        \left( \indij{\ellqk}{\ell} - \pil{k} \right) \indij{\ell}{\ellqtrue}.
    \end{equation}
    We can expand the products on the right and would end up with averages of pair indicators and single ones.
    For the case of pair independent binary classifiers this reduces the 3-way label correlation to,
    \begin{equation}
        \corrijl{i}{j}{\ell} = \left( \frac{1}{Q_\ell} \sum_{q=1}^{Q} \indij{\ellqi}{\ell} \indij{\ellqj}{\ell} 
        \indij{\ellqk}{\ell} \indij{\ell}{\ellqtrue} \right) - \pil{i} \pil{j} \pil{k}.
    \end{equation}
    Setting the 3-way correlation for each label to zero would then give us expressions for three way products.
    A set of 4 classifiers would similarly have 4-way correlations. But for our theorem it does not matter since
    we are only looking at decisions of trios. This means that any set of 3 or more binary classifiers that
    have all their pair correlations and 3-way correlations equal to zero,
    \begin{equation}
        \{ \corrijl{i}{j}{\ell} = 0 \}_{\substack{\ell \in \mathcal{L}\\i \neq j}}, 
        \{ \corrijkl{i}{j}{k}{\ell} = 0 \}_{\substack{\ell \in \mathcal{L}\\i \neq j \neq k}}
    \end{equation}
    could be evaluated by repeated application of this theorem.
    All the products needed for the moments of three classifiers are now expressible as the product of their
    individual performance under our assumption of error independence.

    The construction of each of the generating polynomials in the theorem will be demonstrated with the
    decision pattern (\lbla, \lblb, \lbla). Returning to the label partition of the observed count of
    the pattern,
    \begin{equation}
        n(\lbla, \lblb, \lbla) = n(\lbla, \lblb, \lbla \svbar \lbla) +  n(\lbla, \lblb, \lbla \svbar \lblb),
    \end{equation}
    each term on the right can be rewritten using our product identities and the relation between correct and
    incorrect indicators. Starting with label \lbla,
    \begin{align}
        n(\lbla, \lblb, \lbla \svbar \lbla) & = \sum_{q=1}^{Q} \indij{\ellqi}{\lbla} \indij{\ellqj}{\lblb} 
                                                                \indij{\ellqk}{\lbla} \indij{\ellqtrue}{\lbla}\\
                                            & = \sum_{q=1}^{Q} \indij{\ellqi}{\lbla} (1 - \indij{\ellqj}{\lbla}) 
                                                                \indij{\ellqk}{\lbla} \indij{\ellqtrue}{\lbla}\\
                                            & = Q_\lbla \pia{i} \pia{k} - Q_\lbla \pia{i} \pia{j} \pia{k}\\
                                            & = Q_\lbla \pia{i}(1 - \pia{j}) \pia{k}.
    \end{align}
     The derivation of the label \lblb term is instructive side-by-side with the label \lbla derivation as we can see
     how the second equality in the derivation depends on the true label.
    \begin{align}
        n(\lbla, \lblb, \lbla \svbar \lblb) & = \sum_{q=1}^{Q} \indij{\ellqi}{\lbla} \indij{\ellqj}{\lblb} 
                                                                \indij{\ellqk}{\lbla} \indij{\ellqtrue}{\lblb}\\
                                            & = \sum_{q=1}^{Q} (1 - \indij{\ellqi}{\lblb}) \indij{\ellqj}{\lblb}
                                                                (1 - \indij{\ellqk}{\lblb}) \indij{\ellqtrue}{\lblb}\\
                                            & = Q_\lblb \pib{j} - Q_\lblb \pib{j} \pib{k} 
                                                 - Q_\lblb \pib{i} \pib{j} + Q_\lblb \pib{i} \pib{j} \pib{k} \\
                                            & = Q_\lblb (1 - \pib{i}) \pib{j} (1 - \pib{k}).
    \end{align}

    Putting the terms for each label together and dividing by the size of the test, $Q$, gives
    the desired expression for the observed percentage frequency of the decision pattern
    (\lbla, \lblb, \lbla),
    \begin{align}
        \frac{n(\lbla, \lblb, \lbla)}{Q} & = \frac{Q_\lbla}{Q} \pia{i}(1 - \pia{j}) \pia{k} +  
        \frac{Q_\lblb}{Q} (1 - \pib{i}) \pib{j} (1 - \pib{k})\\
        \fliljlk{\lbla}{\lblb}{\lbla}  &=  \pa \pia{i} (1 - \pia{j}) \pia{k} + \pb (1 - \pib{i}) \pib{j} (1 - \pib{k}).
    \end{align}
    Similar manipulations with the other seven decision patterns concludes the proof of the theorem.
\end{proof}
\label{thm:ind-generating-set}
\end{theorem}

The counterpart to the just proven ``encode'' theorem is the ``decode'' theorem - what are the
statistics of correctness that produced the observed decision tuple frequencies? The encoding
theorem was easy to state and requires only widespread knowledge of algebraic operations.
The decoding theorem requires that one use algebraic geometry.

\begin{theorem}
\label{thm:two-point-ideal}
    Given the frequencies of the eight decision patterns, $\{f_{\ell_i, \ell_j, \ell_k}\}$, for three
    error independent binary classifiers, the algebraic ideal defined by the generating set
    in Theorem~\ref{thm:ind-generating-set} has a variety consisting of two points, one of which
    is the true evaluation.
    \begin{proof}
        Solutions of polynomial systems of equations were the initial impetus for developing
        the mathematical field of algebraic geometry. Given a system of equations (an algebraic
        ideal), what are the set of points (variety) that satisfy all the polynomial equations
        simultaneously? The computational steps needed to derive a closed formula for the
        two points in the theorem are too complex to show on paper. A fully worked out
        notebook (S?) contains the full derivation using the Wolfram language.

        The solution of the ideal for error independent classifiers is carried out by
        creating an alternative representation of the polynomials. Similarly to linear
        systems of equations, we can carry out algebraic transformations of the ideal that
        guarantee the new representation will have the same variety. The strategy of
        generic polynomial solvers, such as the \textbf{Solve} function
        in Wolfram, is to create an \emph{elimination ideal}. This is a representation
        where we can isolate one of the polynomial variables into its own equation.
        Solving this equation we proceed to repeatedly reduce the other equations,
        find polynomials with a single variable, solve and reduce further until all
        solutions have been found. Details of the mathematics of computational algebraic
        geometry needed to prove this theorem can be found in Cox et al.'s textbook\cite{Cox}.

        The variety for the generating set defined by the observable frequencies is a set
        of points in a 7 dimensional space,
        \begin{equation}
            (\pa, \pia{i}, \pib{i}, \pia{i}, \pib{i}, \pia{i}, \pib{i}).
        \end{equation}
        This makes the arbitrary choice of reducing the dimension for the variety
        by applying the relation,
        \begin{equation}
            \pa + \pb = 1
        \end{equation}
        Isolating a variable in this space for a solution, if it exists, can be
        accomplished by establishing a variable order. Putting the prevalence
        at the end of the order produces the elimination ladder we discussed above
        with the \lbla prevalence, \pa, satisfying a quadratic equation,
        \begin{equation}
            \mathcal{P}_2(f_{\ldots}, \ldots) \pa^2 +  
            \mathcal{P}_1(f_{\ldots}, \ldots) \pa +  \mathcal{P}_0(f_{\ldots}, \ldots) = 0.
            \label{eq:label-prevalence}
        \end{equation}
        The coefficients $\mathcal{P}_2, \mathcal{P}_1, \mathcal{P}_0$ are complicated
        polynomials of the observable frequencies $\{ f_{\ell_i, \ell_j, \ell_k}\}.$ In the
        case of the experiment in Table~\ref{tab:by-true-label-partition}, the observed frequencies,
        \begin{equation}
            (\fliljlk{\lbla}{\lbla}{\lbla}, \ldots) = (\frac{568}{20000},
            \frac{553}{20000}, \frac{649}{20000}, \frac{1813}{20000},
            \frac{3534}{20000}, \frac{3607}{20000}, \frac{1068}{20000}, \frac{8208}{20000}),
        \end{equation}
        resulted in the polynomial,
        \begin{equation}
            \frac{3190087950361}{16000000000000000000} \pa^2 -  \frac{3190087950361}{16000000000000000000} \pa +  \frac{1612380721606215379}{1000000000000000000000000}.
        \end{equation}
        A quadratic like Eq~\ref{eq:label-prevalence} has, in general,
        two solutions. As detailed in the notebook showing the full computation (Code S1),
        the elimination ideal that contains this quadratic also contains a subset of linear
        equations relating the prevalence to the unknown label accuracies. These are the
        equations detailed in the Methods and Materials section. As a consequence, once
        one chooses one of the quadratic solutions, all label accuracies are determined by
        linear equations that have a single point solution. We deduce that there only two points 
        in the variety of the ideal given observables. 
        
        By the construction of Theorem~\ref{thm:ind-generating-set}
        we know that one of these must be the true evaluation. The ambiguity in the possible
        evaluations follows from the invariance of polynomials of the generating set in 
        Theorem~\ref{thm:ind-generating-set} to the set of transformations,
        \begin{align}
            \pa & \rightarrow \pb \\
            \pb & \rightarrow \pa \\
            \pia{i} & \rightarrow (1 - \pib{i}) \\
            \pib{i} & \rightarrow (1 - \pia{i}).
        \end{align}
        It follows that error-independent jurors have the smallest possible
        variety and no error correlated ensemble can do better.
    \end{proof}
    
\end{theorem}
    
The last theorem in this Supplement Text deals with the case when we have observed
the decisions of error correlated classifiers. This is the typical case in actual realizations
of binary classifiers. The best we could do is to engineer nearly error-independent classifiers.
What happens when we use the independent generating set for observable frequencies created by
error correlated classifiers? This is an important question in the context of any
evaluation algorithm in unsupervised settings where the evaluations may be critical components
for the safety of a system. How can we detect that the primary assumption in AE,
that the classifiers are error independent on the test, is actually true or not? The appearance
of unresolved square roots - irrational number estimates for the prevalence - is one
way to detect the violation of the assumption as the following theorem states.

\begin{theorem}
    The appearance of irrational roots for the \lbla prevalence quadratic,
    \begin{equation}
            \mathcal{P}_2(f_{\ldots}, \ldots) \pa^2 +  
            \mathcal{P}_1(f_{\ldots}, \ldots) \pa +  \mathcal{P}_0(f_{\ldots}, \ldots).
            \label{eq:label-quadratic}
    \end{equation}
    Can only happen if the classifiers have non-zero error correlations on the test.
    Rational root solutions to the quadratic do not imply that the classifiers are
    error-independent.
    \begin{proof}
        The first part of the theorem is equivalent to saying that the solution for
        error independent classifiers will always be a pair of rationals. This follows
        not from algebraic geometry but from the setting of a finite test where percentage
        prevalences can only be rational numbers (expressible as ratios of integers).
        By Theorem~\ref{thm:two-point-ideal}, we know that error independent will return
        one solution equivalent to its true solution and since this must be a rational,
        the other solution must be too.

        The appearance of irrational solutions for error-correlated classifiers can be
        studied with the full generating ideal. The technical details for this can be
        found here\cite{aeStreaming2023}. 
        It is shown there that the appearance of non-zero correlations prevent
        the factorization of the discriminant of the prevalence quadratic into general terms with
        even powers. For most error correlated trios this means we will observe an irrational
        pair of estimates for the label prevalence. This was, in fact, our observations in all
        of our experiments.
    \end{proof}
\end{theorem}
This theorem highlights the safety nature of the AE. It can tell us that its main assumption
is violated. Caution, however, is necessary in practical applications of this theorem. One
advantage is that there are different ways we can categorize the failures of the estimate
from the AE. Numbers outside the bounds of zero and 1 become possible and, if the classifiers,
are sufficiently correlated, complex estimates are possible. The engineering utility of these
different failure signatures is not resolved by these purely mathematical considerations.
In addition, the error-correlated trios that return rational estimates for the prevalence are
related to random performance and thus an important case to investigate in safety applications.

\subsection*{Related Work}

There are many probabilistic treatments of unsupervised evaluation of classifiers in the
Machine Learning literature. The idea of using the jurors to evaluate themselves seems to have
originated with the work of Dawid and Skene\cite{Dawid79}. Around 2010 two different
approaches continued using probability to understand how to evaluate experts. The first
was Bayesian and exemplified by the work of Raykar et al.\cite{Raykar2010} and a convergence
proof by Jordan's student. The second approach was spectral and started by Parisi et al.
It culminated in the proof by Nadler and collaborators of a $\delta-\epsilon$ solution.
Their work is the probabilistic version of the work done here. Theirs is concerned with
classifiers that make decisions based on factorizable distributions. AE is concerned
with sample statistics of tests.
The use of probability assumptions is not necessary to carry out evaluations as this Report
demonstrates. Empirical and probabilistic approaches should be viewed as complementary.

An algebraic approach to unsupervised evaluation and its connection to logic was
pioneered by Platanios et al.\cite{Platanios2014, Platanios2016} with the
\emph{agreement equations}. The use of logic and algebra alone to derive the
agreement equations points to a logical foundation for unsupervised evaluation.
It can be shown that the agreement equations are a subset
of the complete set needed to generate the frequencies of decision tuples by 
arbitrarily correlated classifiers\cite{aeStreaming2023}. This completeness
allows one to formally verify group evaluations and develop logical alarms
for misaligned classifiers\cite{aeLogical2024}. The independent solution
presented here corrects the one derived using the agreement equations\cite{Platanios2014}
(details in \cite{aeStreaming2023}).

The use of Algebraic Geometry to elucidate statistical questions was pioneered by
Pistone et al\cite{pistone_algebraic_2000} and goes under the name of Algebraic Statistics. 
It has developed into a modern subfield (e.g. \cite{sullivant_algebraic_2023})
and it is mostly concerned with using Algebraic Geometry to develop inference algorithms 
about probabilistic models.

Algebraic Evaluation is a data streaming algorithm. It uses a compressed data sketch of
an observed stream to estimate other statistics. The subject of
streaming algorithms is traditionally thought to be about using low-memory computations.
But their more important aspect is that they are empirical. They encode no knowledge
of the phenomena that is producing the stream. In that sense, Good-Turing frequency
estimation, a streaming algorithm like HyperLogLog, and AE should be viewed as
part of the same class. An explicit connection between AE and the problem of
estimating unique count with noisy ID systems can be found here\cite{agti_2020}.







\clearpage 

\paragraph{Caption for Code S1.}
\textbf{Derivation of closed expressions for the algebraic evaluation
of error independent binary classifiers.} Code notebook, 
\emph{EvaluationIdealAndVarietyErrorIndependentTrio.nb}, details how
the Groebner basis solution of the generating set for error independent
binary classifiers can be expressed as a system of a quadratic polynomial
in \pa and linear equations relating \pa to classifier label accuracies.
A PDF version of the Wolfram language notebook is also available.

\paragraph{Caption for Code S2.}
\textbf{Code and analysis for the American Community Survey experiments} Code notebook, 
\emph{AmericanCommunitySurveyExperiments.nb}, details the labeling experiments.
A PDF version of the Wolfram language notebook is also available.

\paragraph{Caption for Code S3.}
\textbf{Jury evaluation theorem for 3-label error-independent classifiers.} Code notebook, 
\emph{GeneratingSetThreeErrorIndependent3Labels.nb}, shows how to compute the evaluation
for 3-label classifiers.
A PDF version of the Wolfram language notebook is also available.

\paragraph{Caption for Data S1.}
\textbf{Label decisions of four classifiers on 20K ACMEmployment datasets,
$\pa = 1/10.$} Data file, \emph{pa0p1byTrueLabelTestSketches.json}, contains
the label decisions, by true-label, of 1000 labeling experiments with random
draws of 20K from the testing set of about 1M. Each labeling experiment is
summarized as a JSON dictionary mapping true-label to the ensemble decisions
and their observed counts.

\paragraph{Caption for Data S2.}
\textbf{Label decisions of four classifiers on 20K ACMEmployment datasets,
$\pa = 4/10.$} Data file, \emph{pa0p4byTrueLabelTestSketches.json}, has
the same format as Data S1 but for $\pa=4/10$ experiments.

\paragraph{Caption for Data S3.}
\textbf{Label decisions of four classifiers on 20K ACMEmployment datasets,
$\pa = 6/10.$} Data file, \emph{pa0p6byTrueLabelTestSketches.json}, has
the same format as Data S1 but for $\pa=6/10$ experiments.

\paragraph{Caption for Data S4.}
\textbf{Label decisions of four classifiers on 20K ACMEmployment datasets,
$\pa = 9/10.$} Data file, \emph{pa0p9byTrueLabelTestSketches.json}, has
the same format as Data S1 but for $\pa=9/10$ experiments.



\end{document}